\documentclass{article}

\usepackage{arxiv}

\usepackage[utf8]{inputenc} % allow utf-8 input
\usepackage[T1]{fontenc}    % use 8-bit T1 fonts
\usepackage{hyperref}       % hyperlinks
\usepackage{url}            % simple URL typesetting
\usepackage{booktabs}       % professional-quality tables
\usepackage{amsfonts}       % blackboard math symbols
\usepackage{nicefrac}       % compact symbols for 1/2, etc.
\usepackage{microtype}      % microtypography
\usepackage{lipsum}		% Can be removed after putting your text content
\usepackage{graphicx}
\usepackage{doi}

\def\bng{\bngx}

\font\bngx=bang10

\def\*#1*#2{o\null{#2}{#1}}

\def\sh#1{\setbox0=\hbox{#1}%
     \kern-.02em\copy0\kern-\wd0
     \kern.04em\copy0\kern-\wd0
     \kern-.02em\raise.0433em\box0 }

\usepackage{amsmath}
\usepackage{listings}
\usepackage{xcolor} % Optional, for color support
\definecolor{keywordcolor}{rgb}{0.0, 0.0, 1.0} % Blue for keywords
\definecolor{commentcolor}{rgb}{0.0, 0.5, 0.0} % Green for comments
\definecolor{stringcolor}{rgb}{1.0, 0.0, 0.0} % Red for strings
\title{Performance Evaluation of Tokenizers in Large Language Models for the Assamese Language}

\author{ \href{https://orcid.org/0009-0007-8038-1278}{\includegraphics[scale=0.06]{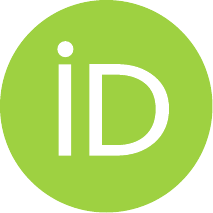}\hspace{1mm}Sagar Tamang}\thanks{
    Correspondance can be addressed to \textit{cs22bcagn033@kazirangauniversity.in}} \\
	Department of IT\\
	The Assam Kaziranga University\\
	Jorhat, India \\
	\texttt{cs22bcagn033@kazirangauniversity.in} \\
	\And
	\href{https://orcid.org/0000-0001-7809-5220}{\includegraphics[scale=0.06]{orcid.pdf}\hspace{1mm}Dr. Dibya Jyoti Bora} \\
	Department of IT\\
	The Assam Kaziranga University\\
	Jorhat, India \\
	\texttt{dibyajyotibora@kazirangauniversity.in} \\
}

\renewcommand{\shorttitle}{Performance Evaluation of Tokenizers in LLMs for the Assamese Language}

\hypersetup{
pdftitle={Performance Evaluation of Tokenizers in LLMs for the Assamese Language},
pdfsubject={cs.AI},
pdfauthor={S. Tamang and D. J. Bora},
pdfkeywords={tokenizer, LLM, tokens, GPT, Assamese, SUTRA},
}

\begin{document}
\maketitle

\begin{abstract}
Training of a tokenizer plays an important role in the performance of deep learning models. This research aims to understand the performance of tokenizers in five state-of-the-art (SOTA) large language models (LLMs) in the Assamese language of India. The research is important to understand the multi-lingual support for a low-resourced language such as Assamese. Our research reveals that the tokenizer of SUTRA from Two AI performs the best with an average Normalized Sequence Length (NSL) value of 0.45, closely followed by the tokenizer of GPT-4o from Open AI with an average NSL value of 0.54, followed by Gemma 2, Meta Llama 3.1, and Mistral Large Instruct 2407 with an average NSL value of 0.82, 1.4, and 1.48 respectively.
\end{abstract}

% keywords can be removed

\section{Introduction}
\label{sec:introduction}
\subsection{Background}
Transformer architecture-based Large Language Models have gained significant popularity in recent years revolutionizing the majority of fields of Artificial Intelligence \cite{vaswani2023attentionneed,islam2023comprehensivesurveyapplicationstransformers}. Tokenization is an important part of the pre-processing step for training and fine-tuning Large Language Models \cite{Toraman_2023}. Thus, the performance of the models also depends on the performance of its tokenizers \cite{Toraman_2023,alrefaie2024exploringtokenizationstrategiesvocabulary}.

Typically, Transformer-based LLMs employ tokenization methods such as WordPiece or Byte Pair Encoding (BPE) \cite{schuster2012,hayase2024}. In the WordPiece method, the primary operation involves counting and merging the most frequent subword pairs. The frequency of a subword pair \((X \to Y)\) can be conceptually represented as \cite{sennrich-etal-2016}:

\begin{equation}
\text{Count}(X \to Y) = \text{number of occurrences of the subword pair } (X, Y) \text{ in the corpus}
\end{equation}

Whereas, in Byte Pair Encoding (BPE) the main operation is to count and merge the most frequent adjacent symbol pairs, and the frequency of an adjacent symbol pair \((a, b)\) can be conceptually determined by \cite{hayase2024}:

\begin{equation}
\text{Count}(ab) = \text{number of occurrences of the pair } (a, b) \text{ in the corpus}
\end{equation}

Byte Pair Encoding (BPE), originally introduced by Sennrich et al. \cite{sennrich-etal-2016} for NLP tasks, is a tokenization algorithm that learns from subword-based encoding from training data. It follows a bottom-up approach where the training dataset is divided into individual characters or tokens, which are then aggregated together in pairs of tokens based on the number of occurrences. BPE has become a near-universal choice for modern language models \cite{hayase2024,sennrich-etal-2016}. 

\subsection{Challenges in Assamese}
Assam is the largest state of the North-Eastern region of India with a population of over 36.6 million. Though the state of Assam harbours a wide range of linguistic groups, Assamese is the official language with 15.1 million people \cite{census2011assam,pathak2022}. We selected Assamese language for our study due to it being a low-resource language, which presents us with a unique challenge and opportunities for evaluating tokenizers. Though there has been research on corpus building, POS tagging, WordNet development, image captioning, and cognate detection, there has been no significant work on Assamese distributional semantics or a monolingual Transformer-based language model for Assamese evaluated on multiple tasks \cite{nath2023,nath2022}.

The rest of the paper is organized as follows: Section \ref{sec:literature} presents an overview of the recent developments in NMT in Assamese language; Section \ref{sec:methodology} provides the methodology of the conduction of this research; Section \ref{sec:experiments} showcases the experiments and its results; Section \ref{sec:discussion} is where the discussions are performed; Section \ref{sec:conclusion} is for the summary of the research with future outlooks.

\section{Literature Review}
\label{sec:literature}
In this section, we conduct a literature review of related works in the related field.

% A. Vaswani et al. reviewed the state-of-the-art NLP models based on Transformers, including BERT, RoBERTa, BART, and DistilBERT. They emphasized the importance of the self-attention mechanism and transfer learning in achieving superior performance across various NLP tasks \cite{vaswani2023attentionneed}.

A. Vaswani et al. (2017) introduced the Transformer model, a groundbreaking architecture for sequence-to-sequence tasks that relies entirely on self-attention mechanisms, eliminating the need for recurrent and convolutional layers. They demonstrated that the Transformer achieves superior performance on machine translation benchmarks, specifically WMT 2014 English-to-German and English-to-French, outperforming previous models. The key innovation of the Transformer is its ability to handle long-range dependencies more effectively, leading to faster training times and improved accuracy in generating complex sequences. Their model has significantly influenced and revolutionized the field of natural language processing \cite{vaswani2023attentionneed}.

\begin{figure}
    \centering
    \includegraphics[width=0.8\linewidth]{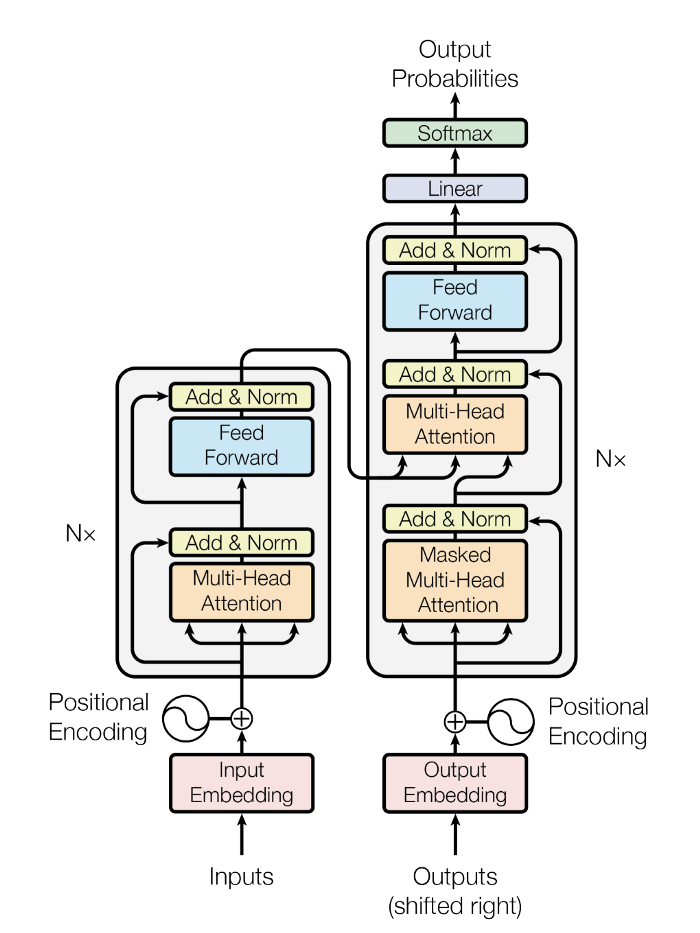}
    \caption{Transformer Architecture from (Vaswani et al., 2017)}
    \label{fig:transformer-architecture}
\end{figure}

G Dagan et al investigated the critical role of tokenization in optimizing large language models (LLMs) for code generation tasks, demonstrating that factors such as tokenizer size, pre-tokenization strategies, and training data significantly influence model performance, generation speed, and memory usage. Through extensive experiments, they revealed that specialized Byte-Pair Encoding (BPE) tokenizers can achieve substantial compression rates without degrading performance, and they provide recommendations for selecting tokenizer hyper-parameters tailored to specific domains \cite{dagan2024}.

\subsection{Tokenization in Low-Resource Languages}
A Bendale et al introduced SUTRA, a scalable multilingual large language model architecture that effectively separates core conceptual understanding from language-specific processing. This innovative design allows SUTRA to achieve high performance and efficiency across over 50 languages, utilizing a Mixture of Experts framework to optimize computational resources and enhance multilingual capabilities, thereby addressing the limitations of existing models and promoting greater accessibility in AI technology \cite{bendale2024}. SUTRA poses as a promising alternative to industry-standard NMT models for the multilingual tasks, specially for the low-resource languages of South-Asia and beyond \cite{sutra2024-deeplearning}.

Conneau et al. (2020) demonstrated that pretraining multilingual language models on a large scale led to substantial performance improvements across a variety of cross-lingual transfer tasks. They trained a Transformer-based masked language model, known as XLM-R, on data from one hundred languages, utilizing over two terabytes of filtered CommonCrawl data. XLM-R achieved significant gains over multilingual BERT (mBERT), including a +14.6\% average accuracy on the XNLI benchmark, a +13\% average F1 score on MLQA, and a +2.4\% F1 score on Named Entity Recognition (NER). The model showed particularly strong performance on low-resource languages, improving XNLI accuracy by 15.7\% for Swahili and 11.4\% for Urdu compared to previous XLM models. They also provided a comprehensive empirical analysis of the factors contributing to these improvements, such as the balance between positive transfer and capacity dilution, and the performance of high and low-resource languages at scale. They highlighted the success of multilingual modeling without compromising per-language performance, with XLM-R being highly competitive against strong monolingual models on the GLUE and XNLI benchmarks. The authors made their code, data, and models publicly available \cite{conneau2020}.

\subsection{Assamese Language}
% A. Nath et al developed a BERT model for Assamese that focuses on masked language modelling (MLM) without sentence prediction (NSP). Their model achieved state-of-the-art results on tokens like Named Entity Recognition and performed well on longer context tasks by leveraging phonological signals \cite{nath-etal-2023-axomiyaberta}.

A. Nath et al developed AxomiyaBERTa, a monolingual BERT-based model tailored for the Assamese language, a low-resource language from Assam, India. They constructed a diverse corpus from Assamese Wikipedia, news articles, and literary texts, which was used to pre-train the model using the masked language modeling (MLM) task, without the next sentence prediction (NSP) task. A key innovation was incorporating Assamese phonological features into the tokenization process, enhancing the model's ability to capture linguistic nuances. Additionally, they introduced an embedding disperser mechanism to address embedding space anisotropy, improving the model's generalization. AxomiyaBERTa outperformed multilingual models like MBERT and XLM in tasks such as Named Entity Recognition (NER), sentiment analysis, and Cloze-style Question Answering (QA), demonstrating its effectiveness in processing Assamese text and setting a precedent for other low-resource languages \cite{nath2023}.

% P. Nath et al. (2022) addressed the critical challenge of the absence of annotated image captioning datasets for the low-resource Assamese language. In their pioneering study, they aimed to develop an image caption generation system tailored specifically for Assamese, a language with limited resources in the field of Natural Language Processing (NLP). Their approach leverages an encoder-decoder framework, which effectively integrates Convolutional Neural Networks (CNNs) with Recurrent Neural Networks (RNNs) to generate meaningful and contextually relevant captions. The researchers conducted experiments using the Flickr30k and COCO Captions datasets, which are originally in English, and translated these datasets into Assamese using advanced Machine Translation (MT) systems. This work not only highlights the significant gap in existing resources but also sets a foundational step towards improving image captioning technology for Assamese, contributing valuable insights into the challenges and solutions in this emerging area of AI and NLP research \cite{nath-etal-2022-image}.

P. Nath et al, highlighting the challenges of the unavailability of annotated datasets for Assamese, conducted a study that aims to develop an image captioning generation for the low-resource Assamese language using an encoder-decoder framework combining Convolutional Neural Network (CNN) and Recurrent Neural Network (RNN) \cite{nath2023}.

% H. Bharali et al. (2014) conducted a comprehensive study on the Assamese language, focusing on the categorization of synonymous words and the analysis of their synonymic patterns. Their research utilized WordNet, a lexical database that organizes words into sets of synonyms called synsets, each defining a unique sense. This study aims to explore and classify the different types of synonymous words in Assamese and to examine their grammatical categories. By leveraging the Assamese WordNet, which contains over 20,000 synsets across nouns, verbs, adverbs, and adjectives, the researchers highlighted various aspects of synonymy, including semantic similarity, connotation, denotation, and stylistic variations. Their work provides valuable insights into the rich and complex nature of synonyms in Assamese, contributing to a deeper understanding of lexical relations and the structure of the language's vocabulary \cite{Bharali2014AnAS}.

H. Bharali et al, working on the Assamese dataset, conducted a study to categorize different types of synonymous words and highlighted their synonymic patterns and grammatical categories found \cite{bharali2014}.

D Pathak et al introduced AsNER, a named entity annotation dataset for Assamese, consisting of approximately 99,000 tokens sourced from the Prime Minister's speeches and Assamese plays. They implemented baseline approaches using state-of-the-art sequence tagging architectures like Bi-LSTM-CRF, achieving the highest F1-score of 80.69\% \cite{pathak2022}.

\subsection{Evaluation Metrics for Tokenizers}

Several works have been conducted to measure the performance of tokenizers in NMT models. For example, F. Qarah and T. Alsanoosy \cite{qarah2024} used the F1 score to gauge performance across multiple metrics for Arabic LLMs. Additionally, they included the Accuracy metric for the majority of the evaluation tasks, whereas the Exact Match (EM) metric was used only for the QA task. EM is a binary evaluation metric that assesses whether the predicted answer exactly matches the referenced answer. The following formulas were used to determine the performance of each classification model:

\begin{equation}
\text{F1-score} = \frac{2 \times \text{Precision} \times \text{Recall}}{\text{Precision} + \text{Recall}} = \frac{2 \times \text{TP}}{2 \times \text{TP} + \text{FP} + \text{FN}}
\label{eq:f1}
\end{equation}

\begin{equation}
\text{Precision} = \frac{\text{TP}}{\text{TP} + \text{FP}}
\label{eq:precision}
\end{equation}

\begin{equation}
\text{Recall} = \frac{\text{TP}}{\text{TP} + \text{FN}}
\label{eq:recall}
\end{equation}

\begin{equation}
\text{Accuracy} = \frac{\text{TP} + \text{TN}}{\text{TP} + \text{TN} + \text{FP} + \text{FN}}
\label{eq:accuracy}
\end{equation}

\begin{equation}
\text{EM} = \frac{\text{Number of Exact Matches}}{T}
\label{eq:em}
\end{equation}

where TP is True Positive, TN is True Negative, FP is False Positive, and FN is False Negative. Additionally, EM represents the average of all individual exact match scores in the set, and \( T \) represents the total number of predictions in the set.

But, for our research, we have used the Vocabulary Size, Average Normalized Sequence Length (NSL) Score, and the Number of Tokens for the analysis of different tokenizers.

\subsection{Gaps in the Literature}
As we have seen through our study, there have been several works done in comparing different tokenizers for arabic LLMs \cite{qarah2024} or on developing LLMs and tokenizers for low-resource languages \cite{bendale2024,sutra2024-deeplearning}, or Assamese \cite{nath2023}, however, there has been no research on comparing or evaluating the performance of different tokenizers of LLMs, specially for the Assamese language. 

\section{Methodology}
\label{sec:methodology}
This section presents how we acquired and analyzed the tokenizers. In this research, we have used the Huggingface platform extensively for its wide availability of ready-made \& open-source models (with tokenizers) \cite{wolf2020}. To be specific, we used the tokenizers from the models listed in Table \ref{table:1} for our study:

\begin{table}[h]
\centering
\caption{Models used in the evaluation along with their Hugging Face details}
\begin{tabular}{lc}
\toprule
\textbf{Name of the Model} & \textbf{Hugging Face Username/Model} \\
\midrule
gpt-4o\_tokenizer & Xenova/gpt-4o \\
google\_gemma\_2\_27b\_it & google/gemma-2-27b-it \\
meta\_llama\_405B & meta-llama/Meta-Llama-3.1-405B \\
mistral\_large\_instruct\_2407 & mistralai/Mistral-Large-Instruct-2407 \\
sutra\_mlt256\_v2 & TWO/sutra-mlt256-v2 \\
\bottomrule
\end{tabular}
\label{table:1}
\end{table}

We used Google Colab with the default CPU hardware accelerator and Python 3 runtime environment for our research. The hugging face transformers' AutoTokenizer function was used to extract the tokenizers from the models listed in the Table \ref{table:1}.

All the tokenizers are used to generate the tokens for an Assamese text to analyze their NSL score and the number of tokens they convert the text into. To maintain consistent results in the experiments, we used an excerpt from an early Assamese novel "Miri Jiyori" by Rajanikanta Bordoloi. \cite{bordoloi2018}. Rajnikanta Bordoloi, was a notable writer, journalist, and tea planter from Assam, India. The novel "Miri Jiyori" unveils some important aspects of then-contemporary Mising society and a series of their customs and traditions. It is a social novel based on a simple love story \cite{miri_jiyori,rajanikanta_bordoloi}.

\begin{quote}
% \textit{Text: {\textassamese{জীৱনৰ পৰিসৰে মোহিত হোৱাটো বাঞ্ছনীয়।}} \\}
\textit{Text: {\bng jiBonar pOriSare mohit hOwaTo baanchaniyo.} \\}
\textit{Translated: It is desirable to be captivated by the expanse of life.} \\
\end{quote}

An important compression metric we used for our analysis of tokenizers, we used the Normalized Sequence Length (NSL), formally defined NSL \( c_{\lambda/\beta} \) as the ratio between the length of an encoded sequence from a tokenizer \( T_\lambda \) and a tokenizer \( T_\beta \). For \( N \) examples taken from a dataset \( D \), the NSL is given by:

\begin{equation}
c_{\lambda/\beta} =
\frac{\sum_{i=1}^N \text{length}(T_\lambda(D_i))}{\sum_{i=1}^N \text{length}(T_\beta(D_i))}
\end{equation}

A Python function for calculating the average Normalized Sequence Length (NSL) and another one for showing the generated tokens were created.

\begin{lstlisting}[language=Python, caption=Python function for calculating NSL]
def calculate_nsl(tokenizer_lambda, tokenizer_beta, dataset):
    nsl_sum_lambda = 0
    nsl_sum_beta = 0
    for data_point in dataset:
        nsl_sum_lambda += len(tokenizer_lambda(data_point))
        nsl_sum_beta += len(tokenizer_beta(data_point))
    return nsl_sum_lambda / nsl_sum_beta
\end{lstlisting}

\begin{lstlisting}[language=Python, caption=Python function for showing tokens]
def show_generated_tokens(tokenizer, dataset):
    for data_point in dataset:
        tokens = tokenizer(data_point)
        print(f"Data point: {data_point}")
        print(f"Tokens: {tokens}")
\end{lstlisting}

\section{Experiments and Results}
\label{sec:experiments}
We have used the Vocabulary Size, Average NSL score, and Number of Tokens as the main data for the analysis of tokens. Typically, higher compression, i.e., less number of tokens or lower average NSL score is considered better but it leads to a higher vocabulary size at the cost of computing and memory \cite{dagan2024}. 

The outputs revealed that among the five models (as seen in Table \ref{table:1}), SUTRA by Two AI's tokenizer \cite{bendale2024} performed the best with an average NSL score and number of Tokens of 0.45 and 16 respectively. Following the SUTRA model's tokenizer closely was the Openai's GPT-4o's Tokenizer with an average NSL score and number of tokens of 0.54 and 19 respectively. Trailing the GPT-4o's tokenizers were the google's Gemma 2's tokenizer \cite{geminiteam2024geminifamilyhighlycapable}, Meta's Llama 405B and the Mistral Large Instruct 2407's tokenizer.

\begin{table}[h]
\centering
\caption{Average NSL and number of tokens for various models (lower better)}
\begin{tabular}{lcccc}
\toprule
\textbf{Name of the Model} & \textbf{Vocab Size} & \textbf{Avg NSL} & \textbf{Number of Tokens} \\
\midrule
gpt-4o\_tokenizer & 200k & 0.54 & 19 \\
google\_gemma\_2\_27b\_it & 256k & 0.82 & 29 \\
meta\_llama\_405B & 128k & 1.4 & 49 \\
mistral\_large\_instruct\_2407 & 32.7k & 1.48 & 52 \\
sutra\_mlt256\_v2 & 256k & 0.45 & 16 \\
\bottomrule
\end{tabular}
\label{table:nsl}
\end{table}

\begin{figure}
    \centering
    \includegraphics[width=0.8\linewidth]{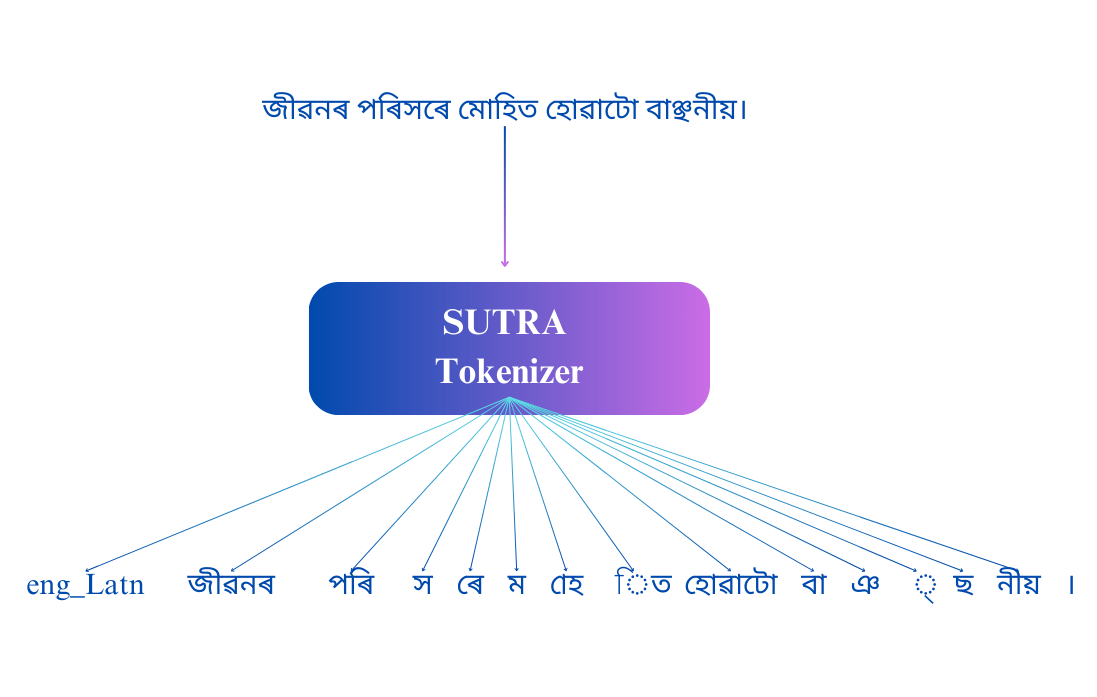}
    \caption{Detailed breakdown of the Example text by the SUTRA Tokenizer.}
    \label{fig:sutra-tokenizer}
\end{figure}

The detailed generation of the tokens can be found in Table \ref{table:tokens}, where we can see the first five tokens generated by the models for the example Assamese text. The detailed split of the Example text by the SUTRA AI is illustrated in the figure \ref{fig:sutra-tokenizer}.

\begin{table*}[h!]
    \centering
    \caption{Breakdown of the example text into Token and Token ID tokens (first five tokens).}
    \begin{tabular*}{\textwidth}{c|c|c|p{10cm}}
        \toprule
        \textbf{Index} & \textbf{Model} & \textbf{Number of Tokens} & \textbf{Breakdown} \\
        \midrule
        1 & gpt-4o & 19 & \begin{tabular}{@{}ll@{}}
                             \textbf{Token} & \textbf{Token ID} \\
                             \midrule
                             à¦ľ & 8271 \\
                             à§Ģà§± & 104175 \\
                             à¦¨à§° & 74976 \\
                             Ġà¦ªà§° & 28325 \\
                             à¦¿à¦¸ & 20919 \\
                             ... & ...\
                             \end{tabular} \\
        \midrule
        2 & Gemma 2 & 29 & \begin{tabular}{@{}ll@{}}
                              \textbf{Token} & \textbf{Token ID} \\
                              \midrule
                              <bos> & 2 \\
                              % \textassamese{জ} & 237642 \\
                              % \textassamese{ী} & 238105 \\
                              % \textassamese{ৱ} & 244307 \\
                              % \textassamese{ন} & 236389 \\
                              \bng j & 237642 \\
                              \bng i & 238105 \\
                              {\bng w} & 244307 \\
                              {\bng n} & 236389 \\
                              ... & ...\\
                              \end{tabular} \\
        \midrule
        3 & Llama 3.1 & 49 & \begin{tabular}{@{}ll@{}}
                               \textbf{Token} & \textbf{Token ID} \\
                               \midrule
                               <|begin\_of\_text|> & 128000 \\
                               à¦ & 11372 \\    
                               ľ & 250 \\
                               à§ & 28025 \\
                               Ģ & 222 \\
                               ... & ... \\
                               \end{tabular} \\
        \midrule
        4 & Mistral Large Instruct 2407 & 52 & \begin{tabular}{@{}ll@{}}
                               \textbf{Token} & \textbf{Token ID} \\
                               \midrule
                               <s> & 1 \\
                               \bng j & 31988 \\
                               \bng i & 32174 \\
                               <0xE0> & 995 \\
                               <0xA7> & 938 \\
                               ... & ... \\
                               \end{tabular} \\
        \midrule
        5 & SUTRA & 16 & \begin{tabular}{@{}ll@{}}
                           \textbf{Token} & \textbf{Token ID} \\
                           \midrule
                           eng\_Latn & 256012 \\
                           \bng jiBonar & 180241 \\
                           \bng pOri & 26897 \\
                           \bng S & 851 \\
                           \bng re & 23288 \\
                           ... & ... \\
                           \end{tabular} \\
        \bottomrule
    \end{tabular*}
    \label{table:tokens}
\end{table*}

\begin{figure}
    \centering
    \includegraphics[width=0.8\linewidth]{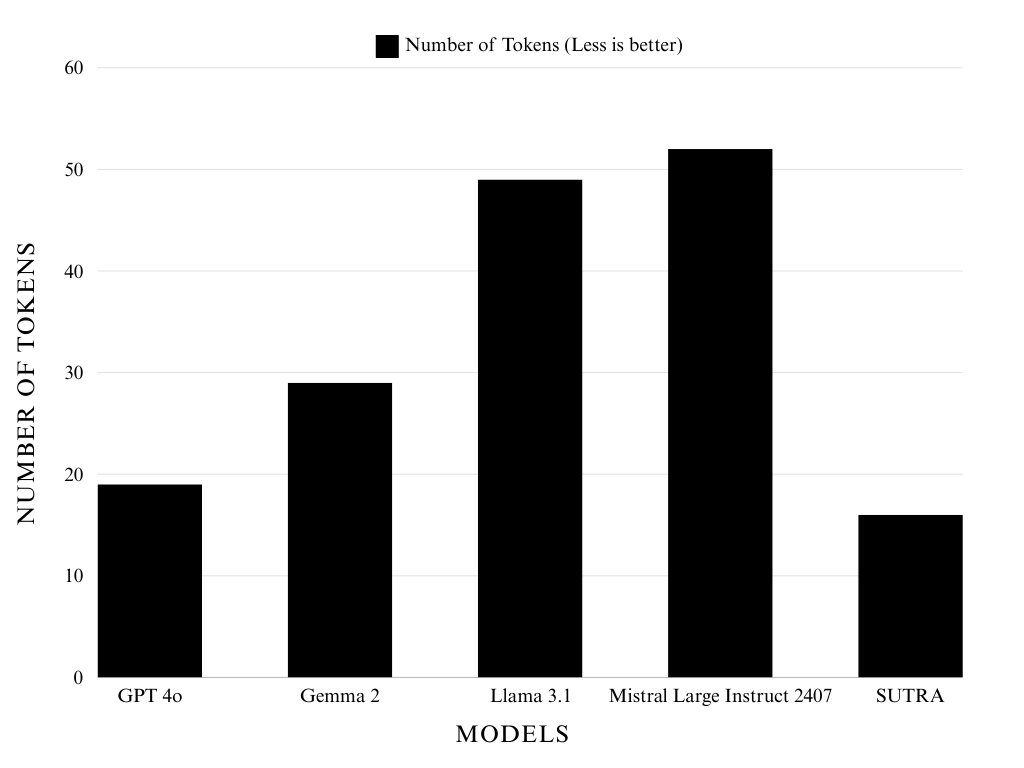}
    \caption{Performance of various LLM Tokenizers on Assamese Text (lower is better)}
    \label{fig:performance-models}
\end{figure}

\section{Discussion}
\label{sec:discussion}
Our findings from the results reveal that among the many models' tokenizers, Sutra's TWO AI performs the best with an average value NSL of 0.45 and generating 16 tokens in the example text. This showcases the multi-lingual capacity of Sutra's TWO AI is well versatile with the Assamese language showcasing its multi-lingual capacity \cite{bendale2024}.

The worst-performing tokenizer for the Assamese language was from the model is Mistral Large Instruct 2407 with an average NSL value of 1.48 and 52 numbers of tokens for our example text. Mistral Large Instruct 2407 is a 123B parameters model with an immense knowledge, but it performed very badly probably due to the limited vocabulary size of just 32.7k, which if compared to the GPT-4o's vocabulary size is 200k (Table \ref{table:nsl}).

Delving deeper in the tokens, in Table \ref{table:tokens}, the tokens generated by GPT-4o, and Llama 3.1 seems to be of different script from the Bengali-Assamese script used by the Assamese language. Whereas, the tokens generated by other models generates the Bengali-Assamese script. This suggests that the GPT-4o and Llama 3.1 tokenizers uses the unicode to work with the Bengali-Assamese script. The tokens generated by Gemma 2, Llama 3.1, and Mistral Large Instruct 2407 shows that it is not able to manage the grouping of words unlike the tokens generated by the GPT-4o or SUTRA. This reveals the lack of presence of Bengali-Assamese script in the tokenizers training dataset. 

\section{Conclusion}
\label{sec:conclusion}
Our study found that the TWO AI's Sutra tokenizers performs the best in Assamese language, followed closely by the tokenizers of GPT-4o, Google's Gemma 2, Llama 3.1, and Mistral Large Instruct 2407, each with an average NSL score of 0.45, 0.54, 0.82, 1.4, and 1.48 respectively (lower the better) as seen in the Table \ref{table:nsl}. However, one area that we could not focus while comparing the tokenizers was that we could not take the vocab size much into account, this can be the future scope of the research. In our research, we have used vocab size, average NSL, and number of tokens to compare the tokenizers, this could further be increased for more matrices to increase the reliability of the verdicts.

Further, similar to the study conducted by F. Qarah and T. Alsanoosy \cite{qarah2024}, where they utilized the F1 score to compare the tokenizers for their performance in Arabic LLMs, and they also trained the LLMs tokenizers on their own by using a dataset, future research can focus on doing something similar to this as well.

We have created a Hugging Face Space titled "\textit{assamese-tokenizer-comparison}" where anyone can compare the performance of different tokenizers on Assamese or other languages. This space is publicly accessible and can be found at: \href{https://huggingface.co/spaces/tamang0000/assamese-tokenizer-comparison}{https://huggingface.co/spaces/tamang0000/assamese-tokenizer-comparison}. 

Hugging Face is a leading platform in the field of Natural Language Processing (NLP), renowned for its open-source library, \texttt{Transformers}, which provides state-of-the-art pre-trained models for a variety of NLP tasks, including text classification, translation, summarization, and question answering. Hugging Face Spaces, specifically, offer a collaborative environment for hosting machine learning demos and experiments, allowing researchers to share models and datasets, reproduce experiments, and contribute to the ongoing development of NLP technologies. By leveraging Hugging Face Spaces, we provide an interactive platform where researchers and developers can test the performance of various tokenizers on Assamese and other low-resource languages.

The "\textit{assamese-tokenizer-comparison}" space allows for direct comparison of tokenizers by providing metrics such as tokenization speed, vocabulary coverage, and sequence length distribution. This initiative aims to facilitate the evaluation and refinement of NLP tools for low-resource languages, which are often underrepresented in NLP research. By providing an easy-to-use interface and leveraging Hugging Face's robust infrastructure, we hope to accelerate research efforts and foster greater inclusion of low-resource languages like Assamese in Neural Machine Translation (NMT) systems. 

Moreover, this space supports the broader mission of Hugging Face to democratize NLP technology and make it accessible to all, regardless of language or resources. We anticipate that this resource will not only promote collaboration within the NLP community but also encourage further exploration and development of language models that are inclusive and representative of diverse linguistic needs.

\section{Acknowledgments}
\label{sec:acknowledgment}
We would like to thank \textit{The Assam Kaziranga University} for providing us with the necessary resources to conduct this research.

%
% ---- Bibliography ----

%Bibliography
% \bibliographystyle{ieeetr}  
% \bibliography{references}
\end{document}